\documentclass[conference]{IEEEtran}
\IEEEoverridecommandlockouts
\usepackage{cite}
\usepackage{amsmath,amssymb,amsfonts}
\usepackage{graphicx, enumitem}
\usepackage{textcomp}
\usepackage[ruled,vlined,linesnumbered]{algorithm2e}
\usepackage[table]{xcolor}
\usepackage{tikz}
\usepackage{bm}
\usepackage{makecell}
\usepackage{booktabs}
\usepackage{multirow}
\usepackage{url}

\newcolumntype{L}[1]{>{\raggedright\let\newline\\\arraybackslash\hspace{0pt}}m{#1}}
\newcolumntype{C}[1]{>{\centering\let\newline\\\arraybackslash\hspace{0pt}}m{#1}}
\newcolumntype{R}[1]{>{\raggedleft\let\newline\\\arraybackslash\hspace{0pt}}m{#1}} 

\def\BibTeX{{\rm B\kern-.05em{\sc i\kern-.025em b}\kern-.08em
    T\kern-.1667em\lower.7ex\hbox{E}\kern-.125emX}}

\begin{document}
	
\title{A Coevolutionary Variable Neighborhood Search Algorithm for Discrete Multitasking (CoVNS): Application to Community Detection over Graphs}

\author{
\IEEEauthorblockN{Eneko Osaba\IEEEauthorrefmark{2}\IEEEauthorrefmark{1},
	Esther Villar-Rodriguez\IEEEauthorrefmark{2}\IEEEauthorrefmark{1} and
	Javier Del Ser\IEEEauthorrefmark{2}\IEEEauthorrefmark{3}}
	\IEEEauthorblockA{\IEEEauthorrefmark{2}TECNALIA, Basque Research and Technology Alliance (BRTA), 48160 Derio, Bizkaia, Spain\\
		Email: [eneko.osaba, esther.villar, javier.delser]@tecnalia.com}
	\IEEEauthorblockA{\IEEEauthorrefmark{3}University of the Basque Country (UPV/EHU), 48013 Bilbao, Bizkaia, Spain\\}
	\IEEEauthorblockA{\IEEEauthorrefmark{1}Corresponding authors. These authors contributed equally to this work.}}
\maketitle

\begin{abstract}
The main goal of the multitasking optimization paradigm is to solve multiple and concurrent optimization tasks in a simultaneous way through a single search process. For attaining promising results, potential complementarities and synergies between tasks are properly exploited, helping each other by virtue of the exchange of genetic material. This paper is focused on Evolutionary Multitasking, which is a perspective for dealing with multitasking optimization scenarios by embracing concepts from Evolutionary Computation. This work contributes to this field by presenting a new multitasking approach named as Coevolutionary Variable Neighborhood Search Algorithm, which finds its inspiration on both the Variable Neighborhood Search metaheuristic and coevolutionary strategies. The second contribution of this paper is the application field, which is the optimal partitioning of graph instances whose connections among nodes are directed and weighted. This paper pioneers on the simultaneous solving of this kind of tasks. Two different multitasking scenarios are considered, each comprising 11 graph instances. Results obtained by our method are compared to those issued by a parallel Variable Neighborhood Search and independent executions of the basic Variable Neighborhood Search. The discussion on such results support our hypothesis that the proposed method is a promising scheme for simultaneous solving community detection problems over graphs.
\end{abstract}

\begin{IEEEkeywords}
Transfer Optimization, Evolutionary Multitasking, Variable Neighborhood Search, Community Detection.
\end{IEEEkeywords}

\section{Introduction} \label{sec:intro}

Transfer Optimization is an incipient research stream within the general field of optimization. Currently, this area is gathering a significant momentum from the related community, leading to an intense scientific production during the last years \cite{ong2016evolutionary}. The main inspiration behind this paradigm is to exploit what has been learned through the optimization of one problem or tasks for the solving of another related or unrelated task. Due to its relatively youth, efforts dedicated to the transferability of knowledge among optimization problems has not been remarkable until recent years, when this concept has become a priority for a wider research community. Arguably, the ever-growing complexity and dimensionality of optimization scenarios has made researchers to turn their attention on methods that allow efficiently harnessing knowledge acquired beforehand.

In this regard, three different categories can be distinguished in Transfer Optimization \cite{gupta2017insights}: sequential transfer \cite{feng2015memes}, multitasking \cite{gupta2016genetic} and multiform optimization. In this paper, we put our attention on the second of these categories. In a nutshell, multitasking is devoted to the simultaneous tackling of different tasks of equal priority by dynamically exploiting existing complementarities and synergies among them. 

More concretely, the present paper is focused on Evolutionary Multitasking (EM, \cite{ong2016towards}), which deals with multitasking optimization scenarios by embracing concepts, operators and search strategies from the area of Evolutionary Computation \cite{back1997handbook,del2019bio}. Related to this specific branch, a particular flavor of EM has shown a remarkable performance when dealing with multitasking environments: Multifactorial Optimization strategy (MFO, \cite{gupta2015multifactorial}). Until now, MFO has been successfully adopted for solving different continuous, discrete, multi- and single-objective optimization tasks \cite{wang2019evolutionary,gong2019evolutionary,yu2019multifactorial,gupta2016multiobjective}. Furthermore, a specific method has garnered most of the literature around this concept: the Multifactorial Evolutionary Algorithm (MFEA, \cite{gupta2015multifactorial}). Unfortunately, alternative methods that populate the EM community are still scarce.

This lack of competitive EM methods is one of the main motivations for the development of this research work. Specifically, this paper proposes a novel EM metaheuristic algorithm based on the well-known Variable Neighborhood Search (VNS, \cite{mladenovic1997variable}) for solving discrete multitasking environments. The Coevolutionary Variable Neighborhood Search Algorithm (CoVNS) herein presented takes a step further beyond the state of the art in two different directions. Firstly, we contribute to the EM field by proposing a new competitive algorithm which, unlike most works published so far in this specific topic, does not hinge on the MFO paradigm. Secondly, CoVNS is a pioneering attempt at exploring the applicability of VNS to the Transfer Optimization paradigm.

Besides the novelty of the method itself, a second contribution of this work relates to the application scenario to which it is applied. It is relevant to first underscore that we focus on discrete optimization In particular, the problem tackled in this work is the detection of communities in weighted directed graphs \cite{pizzuti2017evolutionary}, namely, the optimal partitioning of graph instances whose connections among nodes are directed and weighted. This scenario has been less addressed in the literature than other networks of simpler nature \cite{leicht2008community,osaba2020community}. This being said, to the best of our knowledge this study is the first of its kind dealing with multitasking for solving several community detection problems at the same time. To this end, the discovery of optimal partitions is formulated as an optimization problem, which is driven by a measure of modularity adapted to the directional and weighted nature of the edges of the network \cite{newman2004finding,newman2004analysis}. Results from an extensive experimental setup are presented and discussed to show that the proposed CoVNS excels at solving such multitasking scenarios, outperforming non-multitasking variants of the same algorithm and, hence, providing informed evidence of the benefits of knowledge exchange among tasks.

The remainder of the article is organized as follows. Section \ref{sec:back} provides background and related work. Section \ref{sec:Problem} poses the mathematical formulation of the community detection problems in weighted directed networks. Next, Section \ref{sec:COVNS} exposes in detail the main features of the proposed CoVNS. The experimentation setup and discussion of the results are given in Section \ref{sec:exp}. Finally, Section \ref{sec:conc} concludes the paper with an outlook towards further research.

\section{Background} \label{sec:back}

In order to contextualize this work and properly assess its scientific contribution, this section provides a short overview of the EM research area. In recent years, this scientific branch has emerged as a competitive paradigm for tackling simultaneous optimization tasks. The adoption of evolutionary computation concepts to multitasking (giving rise to EM) has become the \emph{de facto} search strategy: by designing a unified search space, these population-based algorithms allow for an inherent parallel evolution of the whole set of tasks, and for the transfer of genetic material among individuals to exploit inter-task synergies \cite{ong2016evolutionary,gupta2015multifactorial}.

There is a solid consensus that EM was only materialized through the perspective of MFO until late 2017 \cite{da2017evolutionary}. Since then, this incipient research field is gathering a notable corpus of literature focused on new algorithmic schemes, such as the multitasking multi-swarm optimization introduced in \cite{song2019multitasking}, the coevolutionary multitasking scheme proposed in \cite{cheng2017coevolutionary} or the coevolutionary bat algorithm detailed in \cite{osaba2020coeba}. Further alternatives to MFEA have also emerged, partly inspired by the concepts of this influential method. Some examples are the multifactorial differential evolution proposed in \cite{feng2017empirical}, the multifactorial cellular genetic algorithm in \cite{osaba2020multifactorial}, the particle swarm optimization-firefly hybridization introduced in  \cite{xiao2019multifactorial}, or the multifactorial brain storm optimization algorithm presented in \cite{zheng2016multifactorial}. Although in this work the EM environment under consideration is not addressed by using the MFO strategy, we refer interested readers to \cite{bali2019multifactorial,yi2020multifactorial,zhou2020toward} for a recent overview on these methods.

We can mathematically formulate an EM scenario as an environment comprised by $K$ concurrent problems or tasks $T_k$, which must be simultaneously optimized. Thus, the scenario could be characterized by the existence of as many search spaces as tasks. Furthermore, each of the $K$ problems to be solved has a fitness function (objective) $f^k : \Omega_k \rightarrow \mathbb{R}$, where $\Omega_k$ denotes the search space of task $T_k$. We define the main objective of EM as the discovery of a group of solutions $\{\mathbf{x}^{1,\ast},\dots,\mathbf{x}^{K,\ast}\}$ such that $\mathbf{x}^{k,\ast} = \arg \max_{\mathbf{x}\in\Omega_k} f^k(\mathbf{x})$.

An aspect of paramount importance for adequately understanding the above formulation and the EM paradigm itself is that each solution $\mathbf{x}_p$ in the population $\mathbf{P}={\mathbf{x}_p}_{p=1}^P$ is evolved over an unified search space $\Omega^U$, which relates $\Omega_1$ to $\Omega_K$ via an encoding/decoding function $\xi_k: \Omega_k\mapsto \Omega^U$. For this reason, each individual $\mathbf{x}_p\in \Omega^U$ in $\mathbf{P}$ should be decoded to yield a task-specific solution $\mathbf{x}_{p}^k$ for each of the $K$ tasks. 

\section{Problem Statement}\label{sec:Problem}

We now proceed by defining the community detection problem over weighted graphs. First, we model the network as a graph $\mathcal{G}\doteq\{\mathcal{V},\mathcal{E},f_\mathcal{W}\}$, where $\mathcal{V}$ represents the group of $|\mathcal{V}|=V$ nodes or vertices of the network, $\mathcal{E}$ stands for the set of edges connecting every pair of vertices, and $f_\mathcal{W}: \mathcal{V} \times \mathcal{V} \mapsto \mathbb{R}^+$ is a function assigning a non-negative weight to each edge. Furthermore, we consider that $f_\mathcal{W}(v,v)=0$ (i.e. no self loops), and that $f_\mathcal{W}(v,v')=0$ if nodes $v$ and $v'$ are not linked. For notation purposes we define $f_\mathcal{W}(v,v') \doteq w_{v,v'}$, yielding a $V\times V$ adjacency matrix $\mathbf{W}$ given by $\mathbf{W}\doteq \{w_{v,v'}: v,v'\in\mathcal{V}\}$ and fulfilling $\text{Tr}(\mathbf{W})=0$, with $\text{Tr}(\cdot)$ denoting trace of a matrix. Lastly, the directed characteristic of the graph is guaranteed by not imposing any requirement on the symmetry of the adjacency matrix, that is, $w_{v,v'}$ is not necessarily equal to $w_{v',v}$ for any $v\neq v'$.

Using this notation, the task of detecting communities in a network $\mathcal{G}$ can be defined as the partition of the vertex set $\mathcal{V}$ into a number of disjoint, arbitrarily-sized, non-empty groups. Let us denote $M$ as the amount of partitions $\widetilde{\mathcal{V}}\doteq\{\mathcal{V}_1,\ldots,\mathcal{V}_M\}$, such that $\cup_{m=1}^M \mathcal{V}_m = \mathcal{V}$ and $\mathcal{V}_m\cap \mathcal{V}_{m'}=\emptyset$ $\forall m'\neq m$ (i.e., no overlapping communities). Under this formulation, the community to which node $v\in\mathcal{V}$ belongs can be represented as $\mathcal{V}^v\in \widetilde{\mathcal{V}}$.

With all this, we should bear in mind that the weighted directed feature of the graphs used in this paper enforces the reformulation of the in-degree and out-degree values that participate in conventional modularity formulations. A way to redefine such measures is to formulate the so-called input and output \emph{strengths} of node $v$, which are given by:
\begin{equation}
s_v^{in} = \sum_{v'\in\mathcal{V}} w_{v',v}, \qquad s_v^{out} = \sum_{v'\in\mathcal{V}} w_{v,v'},
\end{equation}
that is, as the sum of the weight of the incident (outgoing) edges to (from) node $v$. It is worth noting here that these values represent both the directivity and the weighted nature of adjacency matrix $\mathbf{W}$. Therefore, these two quantities are of paramount importance for properly redefining the concept of communities, in an analogous way to the role played by in- and out-degree values when clustering undirected, unweighted networks.

Bearing all the above formulation in mind, a quality measure for a given partition $\widetilde{\mathcal{V}}$ can be furnished from the main definition of the classical \textit{modularity} for undirected graphs introduced in \cite{newman2004analysis,leicht2008community}. By defining a binary function $\delta: \mathcal{V}\times \mathcal{V} \mapsto \{0,1\}$, so that $\delta(v,v')=1$ if \smash{$\mathcal{V}^v = \mathcal{V}^{v'}$} as per the partition set by \smash{$\widetilde{\mathcal{V}}$} (and $0$ otherwise), the modularity in weighted directed networks can be calculated as:
\begin{equation}
Q(\widetilde{\mathcal{V}}) \doteq \frac{1}{|\sum_\mathbf{W}|}\sum_{v\in\mathcal{V}}\sum_{v'\in\mathcal{V}}\left[w_{v,v'}\frac{s_v^{in}s_{v'}^{out}}{|\sum_{\mathbf{W}}|}\right]\delta(v,v'), \label{Func}
\end{equation} 
where \smash{$|\sum_{\mathbf{W}}|$} represents the sum of the weights of every edge of the graph \cite{chakraborty2017metrics}. Thus, detecting a \emph{high-quality} partition $\widetilde{\mathcal{V}}^\ast$ of a weighted directed network $\mathcal{G}$ can be defined as:
\begin{equation}
\widetilde{\mathcal{V}}^\ast = \arg \max_{\widetilde{\mathcal{V}}\in \mathcal{B}_{V}} Q(\widetilde{\mathcal{V}}),
\vspace{-1mm}
\end{equation}
where $\mathcal{B}_{V}$ stands for the whole set of possible partitions of $V$ elements into nonempty subsets. It is interesting to point out that the cardinality of this set is given by the $V$-th Bell number \cite{harris2008combinatorics}). As a brief example, a small graph composed by $V=20$ nodes amounts up to $517.24\cdot 10^{12}$ possible partitions. Assuming now that the computation of the modularity in \eqref{Func} takes just $1$ microsecond, a practitioner would need more than six months to exhaustively evaluate all the possible partitions. This example is illustrative of the convenience of using heuristics and meta-heuristics for efficiently solving this complex combinatorial problem, and the adoption of multi-tasking approaches when solving several instances of the problem at the same time.

\section{Proposed Variable Neighborhood Search for Discrete Multitasking}\label{sec:COVNS}

Inspired by concepts from previous solvers \cite{cheng2017coevolutionary,osaba2020coeba}, one of the remarkable features of the proposed CoVNS is its multi-population nature. Thus, CoVNS comprises a fixed number of subpopulations or \emph{demes} \cite{luque2011parallel}, composed by the same amount of candidates. The number of subpopulations is equal to the number of tasks $K$ to be solved. Furthermore, each of the $K$ demes $\{\mathbf{P}_k\}_{k=1}^K$ is devoted to the optimization of a specific task $T_k$, meaning that individuals belonging to subpopulation $\mathbf{P}_k$ are only evaluated on task $T_k$ as per its objective $f^k(\mathbf{x})$.

The coevolutionary strategy of CoVNS implies the migration of individuals across subpopulations. Therefore, the consideration of an unified representation $\Omega^U$ becomes necessary. To realize this, the same philosophy of MFEA has been adopted. Nonetheless, one of the main innovative feature of CoVNS is that each deme has its partial view (often restricted by the problem size) of the common search space, potentially requiring a size adjustment when different subpopulations share their individuals. 

Let us focus on the community finding problem for exemplifying this noted size adjustment. First, we encode each individual $\mathbf{x}^k_i$ using a label-based representation \cite{hruschka2009survey}. In this way, each solution $\mathbf{x}_p^k$ belonging to a subpopulation $k$ is denoted as a combination of $V$ integers from the range $[1,\ldots,V]$, where $V$ represents the number of edges in the graph. The value of the $v$-th component of $\mathbf{x}_p^k$ represents the cluster label to which node $v$ belongs. For instance, if we assume a network composed by $V=10$ nodes, a possible individual for task $k$ could be $\mathbf{x}_p^k=[1,2,2,3,3,1,1,2,3,3]$. The communities represented by this individual would be $\widetilde{\mathcal{V}}=\{\mathcal{V}_1,\mathcal{V}_2,\mathcal{V}_3\}$, where $\mathcal{V}_1=\{1,6,7\}$, $\mathcal{V}_2=\{2,3,8\}$ and $\mathcal{V}_3=\{4,5,9,10\}$. Furthermore, the use of this encoding strategy requires a repairing procedure to  avoidance of ambiguities in the representation. To this end, we design a similar procedure to the repairing function proposed in \cite{falkenauer1998genetic}: ambiguities such as those present in $\mathbf{x^k_i}=[2,3,3,4,4,2,2,3,4,4]$ and $\mathbf{x^k_i}=[3,4,4,5,5,3,3,4,5,5]$ (representing both the same partition) are solved by standardizing the solution to $\mathbf{x^k_i}=[1,2,2,3,3,1,1,2,3,3]$.

Turning our attention again to the unified representation $\Omega^U$ used in CoVNS, we denote the dimension of each task $T_k$ (i.e. the number of \emph{nodes}) as $D_k$. Thus, once an individual $\mathbf{x}_p^k\in\Omega_k$ is about to be migrated to a deme in which the dimension of the tasks $T_{k'}$ to be optimized is $D_{k'}<D_k$, only the first $D_k$ elements are considered, reducing in this fashion the phenotype of the solution. In the opposite case, i.e. if $D_{k'}>D_k$, the reverse procedure is carried out. In such a case, and taking into account that when a solution \smash{$\mathbf{x}_p^k$} is transferred to another subpopulation it replaces another individual \smash{$\mathbf{x}_{p'}^{k'}$}, all elements from $D_k$ to $D_{k'}$ are introduced in $\mathbf{x}_p^k$ respecting the order as in \smash{$\mathbf{x}_{p'}^{k'}$}.
\begin{algorithm}[h!]
	\DontPrintSemicolon
	Randomly generate $P$ individuals (initial population)\;
	Evaluate each individual for all the $K$ tasks\;
	Arrange $K$ subpopulations (\emph{demes})\;
	Set $it = 0$\;
	\While{termination criterion not met}{
	    Update iteration counter: $it=it+1$\;
		\For{each deme $k$}{
			\For{each individual $\mathbf{x}_p^k$ in the subpopulation}{
				Generate new solution\;
				\texttt{succFun} = rand($CE_1,CE_3,CC_1,CC_3$)\;
				$\mathbf{x}_p^{new,k} \leftarrow \texttt{succFun}(\mathbf{x}_p^k)$\;				
				\If{$f^k(\mathbf{x}_p^{new,k})>f^k(\mathbf{x}_p^k)$}{
					Accept the new solution $f^k(x_p^{new,k}$\;
				}
			}
		}
		\If{$it\hspace{-1.5mm}\mod{}\textrm{migr}=0$}{
		    \For{each deme k}{
		        \For{$j=1,\ldots,migr\_prop$}{
		        $k'=\text{rand}(1,\ldots,k-1,k+1,\ldots,K)$\;
		        Replace the worst solution in deme $k$ by the best solution in deme $k'$\;
            }
		    }
		}
	}
	Return the best individual in $\mathbf{P}$ for each task $T_k$\;
	\caption{Proposed CoVNS multitasking solver}
	\label{alg:COVNS}
\end{algorithm}

With all this, Algorithm \ref{alg:COVNS} shows the pseudo-code of the proposed CoVNS. As can be seen in this high-level description, in the initialization phase $P$ individuals are randomly generated. Then, each solution is assessed over all the considered $K$ tasks. After this evaluation phase, each subpopulation is generated by choosing the best $P/K$ individuals for the task at hand. This means that the same solution can be chosen for being part of different demes. Once all subpopulations are built, each evolves independently by following the main concepts of a basic discrete VNS. More concisely, each individual, at each iteration, undergoes a successor generation procedure by applying a movement operator on a random basis ($CE_1$, $CE_3$, $CC_1$ or $CC_3$). These operators have been introduced in previous studies \cite{osaba2020community}. For each of these functions, the subscript indicates the amount of randomly chosen nodes, which are extracted from its assigned community. In $CE_\ast$, the chosen elements are re-inserted in already existing communities, whereas in $CC_\ast$ they can be also introduced in newly generated partitions.

Furthermore, every $migr$ iterations, each deme transfers $migr\_prop$ number of individuals to a randomly chosen subpopulation. It should be pointed here that $migr$ = $E \times freq\_migr$, where $E$ represents the number of function evaluations per execution. Furthermore, we set $migr\_prop$ proportional to the population size as $P \times prop$. In our study, and as a result of a thorough empirical process, $freq\_migr = 0.03$ and $prop = 0.05$. Moreover, individuals chosen to be migrated are the $migr\_prop$ best ones, replacing the $migr\_prop$ worst of the destination subpopulation. Lastly, CoVNS completes its search process after $E$ objective function evaluations, after which the best individual of each deme is returned.

\section{Experimental Setup and Results}\label{sec:exp}

For properly gauging the performance of the proposed CoVNS, an extensive set of experiments has been conducted, which is detailed in this section. First, in Section \ref{sec:exp_setup} we elaborate on the benchmark problems used for the proposed algorithm, along with the rest of details of the experimentation setup. Next, in Section \ref{sec:exp_res} we examine and discuss on the results from such experiments.

\subsection{Benchmark Problems and Experimentation Setup}\label{sec:exp_setup}

As has been mentioned in preceding sections, the benefits of the proposed method will be showcased by considering, as tasks, the optimal partitioning of weighted and directed graphs. Accordingly, the performance of CoVNS has been tested over two multitasking scenarios, each composed by $11$ different graph instances. In order to assess the advantage of exchanging genetic material between demes, the performance of our method has been compared to that yielded by two approaches: a separated VNS (sVNS) and a parallel VNS (pVNS). The first approach solves each problem separately by using a single VNS search. For these executions, a fair configuration has been applied for the operators and parameters. The second of the approaches is a parallel implementation of VNS, with no coevolution strategy (each subpopulation evolves independently). Even though no relevant algorithmic differences exist between sVNS and pVNS, the consideration of the parallel approach permits to quantify the contribution of the exchange of knowledge among demes to the convergence of the overall solver.
\begin{table*}[h!]
	\centering
	\vspace{-0.3cm}
	\caption{Parameter values set for CoVNS, pVNS and sVNS.}
	\renewcommand{\arraystretch}{1}
	\resizebox{2.0\columnwidth}{!}{
		\begin{tabular}{lC{2.0cm}C{0.5cm}lC{2.0cm}C{0.5cm}lC{2.0cm}}
			\toprule 
			\multicolumn{2}{c}{CoVNS} & & \multicolumn{2}{c}{pVNS} & & \multicolumn{2}{c}{sVNS} \\
			\cmidrule{1-2} \cmidrule{4-5} \cmidrule{7-8}
			Parameter & Value & & Parameter & Value & & Parameter & Value\\
			\cmidrule{1-2} \cmidrule{4-5} \cmidrule{7-8}
			Population size $P$ & 11$\times$10 & & Population size $P$ & 11$\times$10 & & Population size $P$ & 10 \\ 
			Successor functions & \makecell{$CE_1$, $CE_3$ \\ $CC_1$, $CC_3$} & & Successor functions & \makecell{$CE_1$, $CE_3$ \\ $CC_1$, $CC_3$} & & Successor functions & \begin{tabular}{@{}c@{}}$CE_1$, $CE_3$ \\ $CC_1$, $CC_3$ \end{tabular}\\ 
			Function evaluations & 10$\times$11$\times$1000 & & Function evaluations & 10$\times$11$\times$1000 & & Function evaluations & 10$\times$1000\\
			$freq\_migr$ & 0.03 & &   \\
			$prop$  & 0.05 & &  \\ \bottomrule
		\end{tabular}
	}
	\vspace{-5mm}
	\label{tab:Parametrization}
\end{table*}

Having said that, each multitasking scenario is composed by 11 synthetically generated network instances, which should be optimized in a simultaneous fashion by the three aforementioned methods. Specifically, both benchmarks consist of networks of sizes from 50 to 100 nodes. Each graph has a number of \emph{ground truth} communities, which are modeled by first creating a partition of the network (with random sizes for its constituent communities $\{\mathcal{V}_m\}_{m=1}^M$), and then by connecting nodes within every community with probability $p_{in}$ and nodes of different communities with probability $p_{out}$. Weights $w_{v,v'}$ for every link $(v,v')$ are modeled as uniformly distributed random variables with support $\mathbb{R}[10.0,20.0]$ (intra-community edges) and $\mathbb{R}[0.0,10.0]$ (inter-community edges).

The first environment is called \textit{ordered incremental} (\texttt{OI}), and all the tasks included in this scenario has been named as \texttt{OI\_V\_M}, where $V$ is the number of nodes populating the graph and $M$ the amount of underlying partitions as per the ground truth partition of the network at hand. Regarding $p_{in}$ and $p_{out}$, all datasets have an assigned value of $0.85$ and $0.15$, respectively. The main characteristic of this \texttt{OI} scenario is that instances have been generated in an incremental and ordered way. In other words, new instances have been built by extending the precedent smaller instance respecting the predecessor's graph structure and node identifiers. For instance, all the nodes belonging to the instance \texttt{OI\_60\_8} are also present in the subsequent \texttt{OI\_65\_8} instance, in identical order. Furthermore, the new 5 nodes are added in the 61$^{st}$ to 65$^{th}$ positions of the adjacency matrix. By imposing these conditions our intention is to maintain the order of nodes in the matrix adjacency, guaranteeing that the best solution (partitions) of each instances will share most of their structure.

The second scenario has been coined as \textit{unordered incremental} (\texttt{UI}), naming all the cases as \texttt{UI\_V\_M}, following the same criterion as with the previous \texttt{OI} environment. In these instances, we keep $p_{in} = 0.85$ and $p_{out}=0.15$. Therefore, the main difference between the two environments is that the new incremental nodes in \texttt{UI} are inserted in the first positions, i.e. the new 5 nodes introduced in \texttt{UI\_65\_8} in comparison to \texttt{UI\_60\_8} are added in 1$^{st}$ to 5$^{th}$ positions. This apparently slight modification alters significantly the adjacency matrix and thereby the structure of the best solution corresponding to each incrementally generated graph instance.

The rationale behind this experimental setup follows from influential works \cite{zhou2018study,gupta2016landscape}, which emphasize that one of the most critical aspects when dealing with EM environments is the analysis of the mutual information among the optimized tasks. In fact, it is widely acknowledged that this synergy between tasks is of crucial importance for reaching profitable genetic material exchanges. For this reason, the exploration of what features and characteristics should share different tasks for being synergistic is also valuable in this research context. Therefore, these experiments will help gain a deeper understanding about the conditions that should be met and the performance boundaries when opting for Transfer Optimization in the context of community detection over graphs.

Finally, $20$ independent executions have been carried out for each test case, aiming at shedding light on the statistical significance of eventually discovered performance gapss. Regarding the ending criterion of each method, every run ends after $E = K \times N \times 1000$ objective function evaluations, where $N$ represents the number of individuals per subpopulation. Using this formula, we ensure fairness in comparisons between CoVNS, pVNS and sVNS, dedicating to each approach the same amount of computational resources \cite{LaTorre:20}. To support the replicability of this work, parameters employed for the implemented techniques are shown in Table \ref{tab:Parametrization}.

\subsection{Results and Discussion}\label{sec:exp_res}

Table \ref{tab:results} depicts the results obtained by CoVNS, pVNS and sVNS. Outcomes obtained for each dataset and test case (\texttt{OI} and \texttt{UI}) are given in terms of fitness average, best solution found and standard deviation. It should be mentioned here that the measure used for comparison is the modularity value attained by the solvers (as described in Section \ref{sec:Problem}). In addition, we ease the visualization of the outcomes by highlighting the best average results in bold. Furthermore, in order to ascertain the statistical relevance of differences among algorithms, two different hypothesis tests have been carried out for both \texttt{OI} and \texttt{UI} environments \cite{derrac2011practical}. Results of these tests can be analyzed in Table \ref{tab:resultsStats}. First, the Friedman's non-parametric test for multiple comparison permits proving if differences in performances among the techniques can be cataloged as statistically significant. Thus, first column of Table \ref{tab:resultsStats} depicts the mean ranking returned by this test for each of the compared methods in both test cases (the lower the rank, the better the performance). Furthermore, to assess the statistical significance of the better performance method (CoVNS in both test cases), a Holm's post-hoc test has been performed using our proposal as control solver. This way, the resulting unadjusted and adjusted $p$-values have been included in the second and third columns of Table \ref{tab:resultsStats}. 

Several interesting conclusions can be drawn from Table \ref{tab:results}. To begin with, CoVNS dominates as the best performing method in all the instances that compose the \texttt{OI} multitasking environment. Furthermore, Table \ref{tab:resultsStats} supports the significance of these results at a 99\% confidence level, taking into account that all the $p$-values of the Holm's post-hoc test are lower than $0.01$. These findings statistically conclude that solving \texttt{OI} instances in a simultaneous way and sharing knowledge among different subpopulations contributes to reaching better results. More specifically, since CoVNS has demonstrated to be statistically superior than pVNS, we can confirm that just the simultaneous solving of the tasks is not enough for attaining higher performances. The competitive advantage arises from the efficient sharing of genetic material through individuals belonging to synergistic tasks. As expected, pVNS and sVNS perform similarly, as the only difference between them is the parallelization of the search process (at the level of deme and entire search process, respectively). 

The second important fact is that the structure of the networks is of paramount importance for leveraging genetic transfer. This conclusion becomes evident in the results attained for the \texttt{DI} multitasking environment. In this test case, CoVNS performs best in 6 out of 11 instances, although the overall performance gap is not statistically significant as observed in Table \ref{tab:resultsStats}. These outcomes clearly brings us to the conclusion that the genetic material sharing among non-complementary instances does not provide any competitive advantage for the search process. We recall at this point that, as opposed to \texttt{OI} instances, in \texttt{UI} tasks the structure of incrementally generated graphs changes considerably as more nodes are added to the graphs. Therefore, we conclude that although CoVNS seemingly outperforms both pVNS and sVNS, there is no statistical evidence that the sharing of knowledge leads to significant better outcomes.

In fact, this analysis leads to the two main conclusions of this paper. This first one regards the composition of complementary graphs. As observed in this experimentation, for materializing positive genetic transfer among tasks, network instances should share their structure in an incremental way as explained in the case of \texttt{OI} so as to enforce a degree of overlap between their optimal partitions. Secondly, CoVNS has demonstrated to be a promising method for simultaneous solving community detection problems over graphs, obtaining significant competitive advantages whenever the networks are interrelated.
\begin{table*}[h!]
	\centering
	\caption{Results obtained by CoVNS, pVNS and sVNS for all both test environments. Best average results have been highlighted in bold. Each (algorithm,instance) cell indicates average (top), best (middle) and standard deviation (bottom) of the modularity fitness computed over 20 independent runs.}
	\renewcommand{\arraystretch}{0.92}
	\resizebox{1.95\columnwidth}{!}{
		\begin{tabular}{cccccccccccccc}
			\toprule
			& & \makecell{\texttt{OI\_50\_8}} & \makecell{\texttt{OI\_55\_8}} & \makecell{\texttt{OI\_60\_8}} & \makecell{\texttt{OI\_65\_8}} & \makecell{\texttt{OI\_70\_8}} & \makecell{\texttt{OI\_75\_8}} & \makecell{\texttt{OI\_80\_8}} & \makecell{\texttt{OI\_85\_8}} & \makecell{\texttt{OI\_90\_8}} & \makecell{\texttt{OI\_95\_8}} & \makecell{\texttt{OI\_100\_8}} & \\ \midrule
			\multirow{10}{*}{\rotatebox{90}{\emph{Ordered Incremental}}} & 
			\multirow{3}{*}{CoVNS} 
			& \textbf{0.330} & \textbf{0.322} & \textbf{0.342} & \textbf{0.311} & \textbf{0.291} & \textbf{0.301} & \textbf{0.276} & \textbf{0.256} & \textbf{0.247} & \textbf{0.252} & \textbf{0.230} \\ 
			& & 0.365 & 0.354 & 0.379 & 0.348 & 0.324 & 0.328 & 0.302 & 0.282 & 0.272 & 0.283 & 0.271\\
			& & 0.022 & 0.025 & 0.026 & 0.023 & 0.024 & 0.021 & 0.017 & 0.015 & 0.015 & 0.020 & 0.022\\ 
			\cmidrule{2-13}
			& \multirow{3}{*}{pVNS} 
			& 0.322 & 0.294 & 0.280 & 0.252 & 0.224 & 0.224 & 0.200 & 0.179 & 0.172 & 0.165 & 0.157\\ 
			& & 0.360 & 0.337 & 0.305 & 0.302 & 0.243 & 0.255 & 0.218 & 0.198 & 0.197 & 0.191 & 0.174\\
			& & 0.022 & 0.024 & 0.021 & 0.024 & 0.010 & 0.015 & 0.010 & 0.008 & 0.013 & 0.011 & 0.009\\  
			\cmidrule{2-13}
			& \multirow{3}{*}{sVNS} 
			& 0.319 & 0.286 & 0.290 & 0.260 & 0.229 & 0.226 & 0.205 & 0.189 & 0.169 & 0.172 & 0.160\\
			& & 0.344 & 0.307 & 0.318 & 0.284 & 0.254 & 0.271 & 0.221 & 0.202 & 0.198 & 0.193 & 0.171\\ 
			& & 0.014 & 0.016 & 0.024 & 0.018 & 0.132 & 0.020 & 0.012 & 0.010 & 0.012 & 0.009 & 0.008\\
			\midrule
			& & \makecell{\texttt{UI\_50\_8}} & \makecell{\texttt{UI\_55\_8}} & \makecell{\texttt{UI\_60\_8}} & \makecell{\texttt{UI\_65\_8}} & \makecell{\texttt{UI\_70\_8}} & \makecell{\texttt{UI\_75\_8}} & \makecell{\texttt{UI\_80\_8}} & \makecell{\texttt{UI\_85\_8}} & \makecell{\texttt{UI\_90\_8}} & \makecell{\texttt{UI\_95\_8}} & \makecell{\texttt{UI\_100\_8}} & \\ \midrule
			\multirow{10}{*}{\rotatebox{90}{\emph{Unordered Incremental}}} & 
			\multirow{3}{*}{CoVNS} 
			& 0.299 & 0.279 & \textbf{0.287} & 0.251 & \textbf{0.227} & \textbf{0.231} & \textbf{0.205} & 0.180 & \textbf{0.169} & 0.168 & \textbf{0.164} \\ 
			& & 0.325 & 0.325 & 0.316 & 0.270 & 0.257 & 0.262 & 0.247 & 0.200 & 0.193 & 0.186 & 0.194\\
			& & 0.015 & 0.025 & 0.020 & 0.015 & 0.015 & 0.017 & 0.021 & 0.012 & 0.010 & 0.010 & 0.012\\ 
			\cmidrule{2-13}
			& \multirow{3}{*}{pVNS} 
			& \textbf{0.323} & \textbf{0.282} & 0.270 & 0.243 & 0.226 & 0.222 & 0.201 & 0.183 & 0.167 & \textbf{0.169} & 0.163\\ 
			& & 0.369 & 0.317 & 0.293 & 0.284 & 0.245 & 0.259 & 0.228 & 0.203 & 0.196 & 0.193 & 0.183\\
			& & 0.019 & 0.019 & 0.016 & 0.020 & 0.009 & 0.018 & 0.010 & 0.015 & 0.012 & 0.010 & 0.014\\  
			\cmidrule{2-13}
			& \multirow{3}{*}{sVNS} 
			& 0.322 & 0.295 & 0.280 & \textbf{0.258} & 0.219 & 0.217 & 0.201 & \textbf{0.203} & 0.166 & 0.162 & 0.152\\
			& & 0.375 & 0.340 & 0.317 & 0.299 & 0.270 & 0.250 & 0.223 & 0.224 & 0.189 & 0.177 & 0.177\\ 
			& & 0.027 & 0.025 & 0.019 & 0.021 & 0.021 & 0.019 & 0.013 & 0.011 & 0.012 & 0.010 & 0.012\\
			\bottomrule
		\end{tabular}
	}
	\label{tab:results}
\end{table*}

\begin{table}[htb]
	\centering
	\caption{Results of the Friedman's non-parametric tests, and unadjusted and adjusted $p$-values obtained through the application of Holm's post-hoc procedure using CoVNS as control algorithm.}
	\renewcommand{\arraystretch}{1.25}
	\scalebox{1.0}{
		\begin{tabular}{c c c c c}
		\toprule
			& & Friedman's Test & \multicolumn{2}{c}{Holm's Post Hoc}\\
			\cmidrule{3-5}
			& & Rank & Unadjusted $p$ & Adjusted $p$\\
			\cmidrule{1-5}
			\multirow{3}{*}{\rotatebox{90}{\emph{OI}}} & CoVNS & 1 & -- & --\\
			& pVNS & 2.7273 &0.000051 &0.000102\\
			& sVNS & 2.2727 & 0.002838 & 0.002838\\
			\midrule
			\multirow{3}{*}{\rotatebox{90}{\emph{DI}}} & CoVNS & 1.7273 & -- & --\\
			& pVNS & 2.0455 &0.240955 &0.481909\\
			& sVNS & 2.2273 & 0.455545 & 0.481909\\
			\bottomrule
		\end{tabular}
	}	
	\label{tab:resultsStats}
\end{table}

\section{Conclusions and Future Work}\label{sec:conc}

This paper has elaborated on the design, implementation and validation of a novel Coevolutionary Variable Neighborhood Search algorithm for dealing with evolutionary multitasking scenarios. The proposed method relies on a discrete adaptation of the VNS heuristic, incorporating further elements from co-evolutionary multitasking algorithms \cite{cheng2017coevolutionary,osaba2020coeba}. In addition to the method itself, an equally important contribution of this work is the first attempt at applying Transfer Optimization to community detection over weighted and directed graphs. In this way, we have compared the results attained by CoVNS over two test cases composed of 11 datasets with the ones furnished by a parallel (not coevolutionary) VNS and by independent executions of the VNS. The obtained results validate our hypothesis: the knowledge sharing that lies at the heart of CoVNS is crucial for reaching better results when simultaneously solving complementary tasks.

Several research lines have been arranged as future work. In the short term, we plan to evaluate the scalability of the proposed method by analyzing its computational efficiency when simultaneously dealing with a high number of cases. We will also explore the adaptation of the method to other combinatorial optimization problems stemming from other research fields \cite{precup2019nature}. In a longer term, we plan to endow this method with enhanced adaptive mechanisms so as to automatically define the optimal strategy for sharing knowledge according to the detected level of relationship amongst tasks. To this end, we plan to design schemes for automatically detecting the level synergy of the optimizing graphs during the search process, in order to autonomously boost the transfer of knowledge. We expect that these methods, currently under active investigation in other related works \cite{bai2019simgnn}, will help the solver adaptively harness positive knowledge transfers, and stay resilient against negative (hence, counterproductive) genetic shares.

\section*{Acknowledgments}

The authors would like to thank the Spanish Centro para el Desarrollo Tecnologico Industrial (CDTI, Ministry of Science and Innovation) through the ``Red Cervera'' Programme (AI4ES project), as well as by the Basque Government through EMAITEK and ELKARTEK (ref. 3KIA) funding grants. J. Del Ser also acknowledges funding support from the Department of Education of the Basque Government (Consolidated Research Group MATHMODE, IT1294-19).

\bibliographystyle{IEEEtran}
\bibliography{IEEEexample}

\end{document}